\documentclass[manuscript,screen]{acmart}
\AtBeginDocument{%
  }

\setcopyright{acmlicensed}
\copyrightyear{2018}
\acmYear{2018}
\acmDOI{XXXXXXX.XXXXXXX}
\acmConference[Conference acronym 'XX]{Make sure to enter the correct
  conference title from your rights confirmation email}{June 03--05,
  2018}{Woodstock, NY}
\acmISBN{978-1-4503-XXXX-X/2018/06}

\usepackage{multirow}
\usepackage{tikz}
\usetikzlibrary{positioning,arrows.meta}
\usepackage{wrapfig}
\definecolor{dropYes}{HTML}{C62828}     % 深红
\definecolor{dropSlight}{HTML}{EF6C00}  % 橙
\definecolor{dropMinor}{HTML}{FF8F00}   % 浅黄
\definecolor{dropNo}{HTML}{2E7D32} 
\usepackage{pifont}          % 提供 \ding 命令
\newcommand{\cmark}{\checkmark}
\newcommand{\xmark}{%
  \raisebox{0.65ex}{%
    \rotatebox[origin=c]{45}{\rule{1.2ex}{.15ex}}%
    \kern-0.9ex%
    \rotatebox[origin=c]{-45}{\rule{1.2ex}{.15ex}}}}
\newcommand{\hmark}{%
  \checkmark\kern-1.1ex%
  \raisebox{.6ex}{\rotatebox[origin=c]{125}{--}}}
\begin{document}

\title{Out-of-Distribution Segmentation in Autonomous Driving}
\title{From Pixel to Mask: A Survey of Out-of-Distribution Segmentation}
\author{Wenjie Zhao}
\email{wxz220013@utdallas.edu}
\orcid{1234-5678-9012}
\affiliation{%
  \institution{University of Texas at Dallas}
  \city{Richardson}
  \state{Texas}
  \country{USA}
}

\author{Jia Li}
\email{jxl220096@utdallas.edu}
\orcid{1234-5678-9012}
\affiliation{%
  \institution{University of Texas at Dallas}
  \city{Richardson}
  \state{Texas}
  \country{USA}
}

\author{Yunhui Guo}
\email{yunhui.guo@utdallas.edu}
\orcid{1234-5678-9012}
\affiliation{%
  \institution{University of Texas at Dallas}
  \city{Richardson}
  \state{Texas}
  \country{USA}
}

\renewcommand{\shortauthors}{Zhao et al.}

\begin{abstract}
  % Out-of-distribution (OoD) detection has attracted growing attention as concerns about AI security rise. Although detecting OoD images is valuable, it merely flags their presence without spatial context, limiting task-specific responses.
  % By contrast, OoD segmentation is essential for safety-critical applications such as autonomous driving. A perception module must not only be aware of the presence of OoD objects but also localize their exact shapes and positions, enabling the vehicle to take targeted control actions and thereby enhance overall system robustness. 
  % In this survey, we group current OoD segmentation approaches into four categories: (i) test-time OoD segmentation, (ii) outlier exposure for supervised training, (iii) reconstruction-based methods, (iv) and approaches that leverage powerful models. We review recent progress in autonomous-driving settings and discuss emerging challenges and future directions.
 Out-of-distribution (OoD) detection and segmentation have attracted growing attention as concerns about AI security rise. Conventional OoD detection methods identify the existence of OoD objects but lack spatial localization, limiting their usefulness in downstream tasks.
  OoD segmentation addresses this limitation by localizing anomalous objects at pixel-level granularity. This capability is crucial for safety-critical applications such as autonomous driving, where perception modules must not only detect but also precisely segment OoD objects, enabling targeted control actions and enhancing overall system robustness.
  In this survey, we group current OoD segmentation approaches into four categories: (i) test-time OoD segmentation, (ii) outlier exposure for supervised training, (iii) reconstruction-based methods, (iv) and approaches that leverage powerful models. We systematically review recent advances in OoD segmentation for autonomous-driving scenarios, identify emerging challenges, and discuss promising future research directions.
\end{abstract}





\maketitle

% \section{Introduction}

% \section{Template Overview}

% \subsection{Template Styles}

\section{Introduction}
\label{sec:intro}

Autonomous driving has attracted growing attention due to its potential to free humans from driving. Automated driving systems (ADS) typically comprise multiple modules, including a perception module that gathers and processes environmental data to guide decision-making. The remarkable success of deep neural networks in computer vision tasks has made it feasible to integrate them into the perception module of ADS. While deep neural networks have demonstrated significant success in standard vision benchmarks \cite{alexnet}, deploying them into automated driving systems poses substantially greater challenges compared to controlled environments such as ImageNet \cite{imagenet_challenge}. ADS operates in real-world environments where encountering unexpected scenarios is common. Such scenarios often involve objects that did not occur during training.

Vanilla deep neural networks often exhibit overconfidence when encountering objects not seen during training, leading to incorrect classifications into known categories \cite{blum2019fishyscapes}. Such shortcomings within the perception module introduce unacceptable safety risks for ADS. Ideally, an automated driving system should be robust across diverse conditions and clearly recognize its limitations, alerting the driver or adopting more conservative strategies when encountering scenarios beyond its capabilities \cite{rios2022managing}. To achieve this, the model must accurately identify inputs that it cannot reliably process. In other words, it should effectively detect out-of-distribution data unseen during training. The concept of OoD detection was initially formulated by Hendrycks \cite{hendrycks2016baseline}, in image classification settings and later extended to semantic segmentation, which is crucial for ADS due to its requirement for pixel-level localization of anomalies.

OoD segmentation evolved naturally from the OoD detection problem. Early studies adapted existing OoD detection methods to segmentation tasks and achieved promising results. For example, Rottmann et al. \cite{rottmann2020prediction} proposed an OoD segmentation method by aggregating multiple dispersion measures. Some researchers have also attempted to improve models' sensitivity to OoD data by training on dataset generated by outlier exposure \cite{bevandic2018discriminative, bevandic2019simultaneous, chan2021entropy, tian2022pixel, grcic2022densehybrid, liu2023residual}. Others have explored reconstruction-based methods, identifying anomalies by comparing reconstructed images with the original inputs \cite{7225680, lis2019detecting, xia2020synthesize, haldimann2019not, vojir2021road, vojivr2023image, di2021pixel}. More recently, approaches leveraging powerful pretrained models for OoD segmentation has attracted increasing attention. Despite the rapid growth of research in this area, a comprehensive and systematic review of recent advances is still lacking. To fill this gap and facilitate future research, we present a survey of OoD segmentation approaches specifically in autonomous driving scenario.

\begin{wrapfigure}{r}{0.63\linewidth}  % 'r' = right side; adjust width as needed
  \centering
  \includegraphics[width=\linewidth]{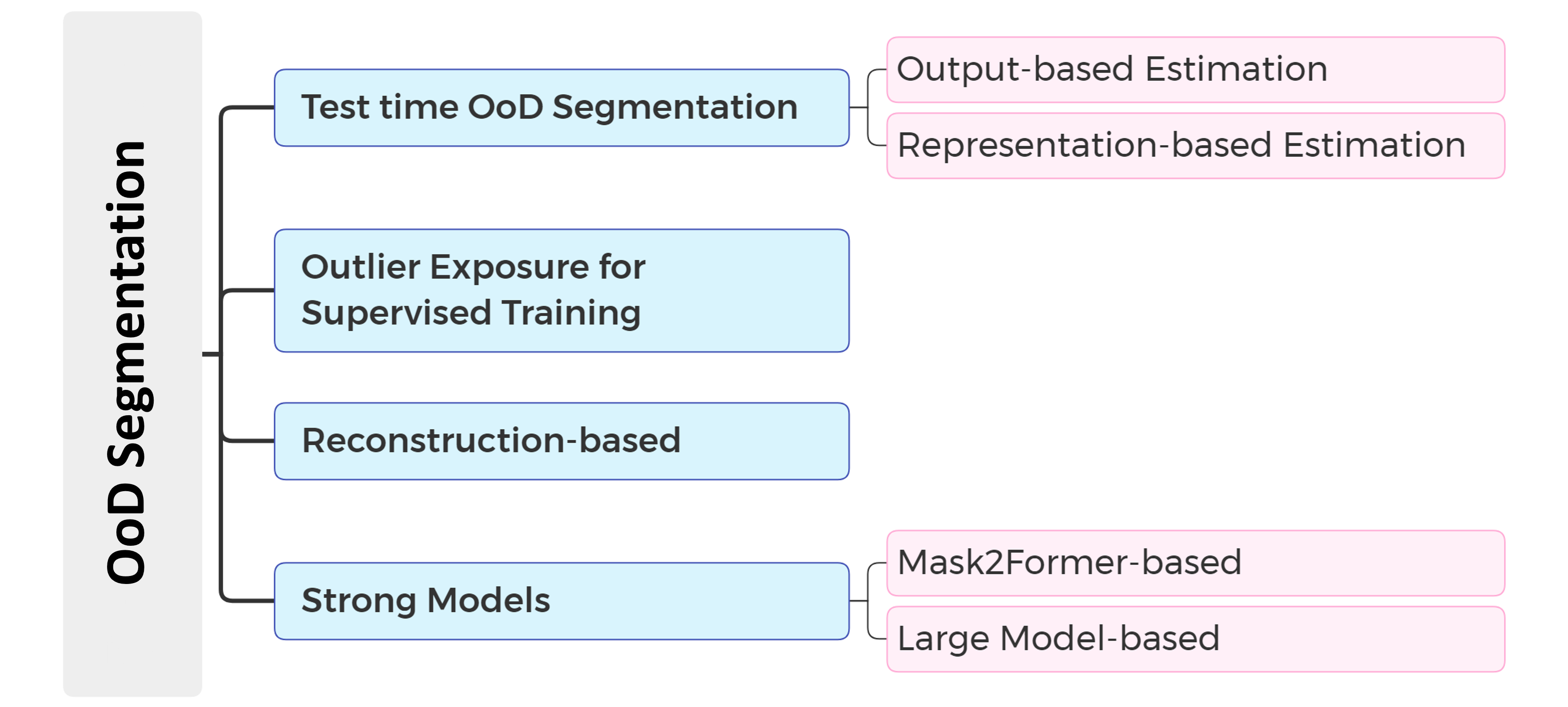}
  \caption{Taxonomy of methods for OoD segmentation in autonomous driving.}
  \label{fig:method}
\end{wrapfigure}

In this survey, we first describe the problem setup for out-of-distribution segmentation in autonomous driving in Section \S\ref{sec:problemsetup}. We then investigate datasets and the metrics in Section \S\ref{sec:dataset}. Furthermore, we review the methods proposed for OoD segmentation in autonomous driving in Section \S\ref{sec:method}, which includes four categories. Finally, we review the potential challenges and outline future directions in the field of OoD segmentation for autonomous driving in Section \S\ref{sec:conclusion}.

\section{Problem Setup/ Problem review}
\label{sec:problemsetup}

Given a labeled training dataset drawn from an in-distribution environment $P_{\text{in}}(X,\mathbf Y)$, each image $X\in\mathbb{R}^{H\times W\times C}$ is paired with a pixel-wise mask $\mathbf Y\in\mathcal{C}_{\text{in}}^{H\times W}$, where
$\mathcal{C}_{\text{in}}=\{1,\dots,K\}$ denotes the set of known classes. A model $f_\theta$ trained on this data processes an input image $X$ and
produces a logits vector
$z(x)\!\in\!\mathbb{R}^{K}$ for every pixel $x\!\in\!X$. These logits are then converted into class probabilities.  Under the closed-set assumption, the model is expected to correctly segment images into the known label space.
In real deployments, however, the
system may encounter inputs from an unknown distribution
$P_{\text{out}}(X,\mathbf Y)$ that contain pixels whose ground-truth
classes belong to a disjoint set $\mathcal{C}_{\text{out}}$ satisfying
$\mathcal{C}_{\text{in}}\cap\mathcal{C}_{\text{out}}=\varnothing$.

Pixel-level OoD segmentation distinguishes ID and OoD regions at the granularity of individual pixels, producing a binary mask $M \in \{0,1\}^{H\times W}$ that labels each pixel as ID (0) or OoD (1):
\[
M_{h,w} = 
\begin{cases}
0, & \text{if pixel } (h,w)\in X \text{ belongs to ID region},\\[4pt]
1, & \text{if pixel } (h,w)\in X \text{ belongs to OoD region}.
\end{cases}
\]

% In order to perform segmentation, the model processes the input image $X$ and produces a logits vector $z(x) \in \mathbb{R}^K$ for each pixel $x \in X$, where $K$ is the total number of known classes. 

% Given pixel-level distributions $P_{\text{in}}(X_{h,w})$ and $P_{\text{out}}(X_{h,w})$, pixel-level OoD segmentation learns a decision function $s_\phi: \mathbb{R}^{H\times W\times C}\rightarrow \{0,1\}^{H\times W}$ parameterized by $\phi$, \textcolor{red}{to accurately segment OoD regions}:
\[
% \min_{\phi}\mathbb{E}_{X_{h,w}\sim P_{\text{in}},P_{\text{out}}}\left[\ell_{\text{seg}}\left(s_\phi(X)_{h,w}, M_{h,w}\right)\right],
\]

\section{Benchmarks}
\label{sec:dataset}
In this section, we review widely adopted datasets for OoD segmentation in autonomous driving. We also introduce commonly used evaluation metrics. 
\subsection{Datasets}

\begin{table}[h]
\centering
\resizebox{\textwidth}{!}{  % 自动缩放宽度到页面宽
\begin{tabular}{lcccc}
\toprule
\textbf{Dataset} & \textbf{Resolution} & \textbf{Val / Test} & \textbf{Data Source} & \textbf{Evaluation Area} \\
\midrule
LostAndFound     & 2048 $\times$ 1024                         & 1036 / 1068     & Captured                & Road-only     \\
RoadAnomaly      & 1024 $\times$ 512                          & 60 / -          & Collected               & Road-context  \\
FS Static        & 2048 $\times$ 1024                         & 30 / 1000       & Synthetic               & Road-context  \\
FS L\&F          & 2048 $\times$ 1024                         & 100 / 275       & Collected               & Road-context  \\
SMIYC Anomaly    & 2048 $\times$ 1024 or 1280 $\times$ 720    & 10 / 100        & Collected               & Road-context  \\
SMIYC Obstacle   & 1920 $\times$ 1080                         & 30 / 327        & Collected and Captured  & Road-only     \\
\bottomrule
\end{tabular}
}
\caption{Summary of representative OoD segmentation datasets in autonomous driving. Each dataset is characterized by its resolution, validation/test split, data source, and evaluation area.}
\label{tab:dataset-comparison}
\end{table}

\begin{wrapfigure}{r}{0.4\linewidth} 
% 'r' = right side; adjust width as needed
  \centering
  \includegraphics[width=\linewidth]{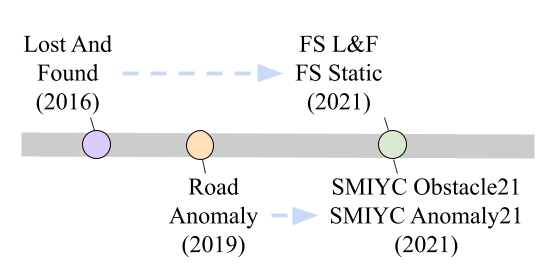}
  \caption{Chronological development of datasets used for OoD segmentation.}
  \label{fig:dataset}
\end{wrapfigure}

\textbf{LostAndFound} was introduced by Pinggera et al. \cite{pinggera2016lost} in 2016. It was the first dataset designed to evaluate small obstacle segmentation, a critical task for autonomous driving safety. It consists of real-world image captured from 13 challenging street scenarios, featuring 37 different obstacle types that vary in size and material. The dataset contains 2,104 images at a resolution of 2048 × 1024 pixels, with 1,036 images allocated for training/validation and 1,068 for testing. Annotations in this dataset focus solely on the road region, including masks for obstacles and free road space. Notably, buildings and roadside structures are not annotated.
% small road hazard, real image and ood, proposed at 19 by pinggera. Only care about road area and obstacle on the road. same set up with cityscape.

\textbf{RoadAnomaly} was proposed by Lis et al. \cite{lis2019detecting} in 2019. It aims to support the detection of unexpected objects in road scenes. It includes 60 real-world images collected online, with anomalous objects such as animals, rocks, tires, trash, cans, and construction equipment. These images have a uniform resolution of 1280 × 720 pixels and were resized to 1024 × 512 for evaluation. Annotations cover both the road surface and its immediate surroundings. Compared to LostAndFound dataset, which focuses solely on small road obstacles within the road area, RoadAnomaly dataset covers anomalies of various scales. Additionally, it extends the region of interest beyond the road to include the roadside context.
% Care road area (include road itself and nearby.)\ real image and ood object \ 1024*512 \cite{lis2019detecting}

\textbf{Fishyscapes} was proposed by Blum et al. \cite{blum2021fishyscapes} in 2021 for anomaly segmentation. It consists of three subsets known as FS Static, FS Web and FS Lost \& Found. FS Static provide a public validation set of 30 images and a private test set of 1,000 images, while FS Lost \& Found offers an accessible validation set of 100 images and a private test set of 275 images. All validation images have a resolution of 2048 × 1024 pixels. \textbf{FS Static} and FS Web share the same construction methodology. The authors overlaid anomalous objects, such as animals and household items, extracted from Pascal VOC \cite{everingham2010pascal} onto background images from the Cityscapes validation set \cite{cordts2016cityscapes}. These anomalous objects are carefully selected to ensure that their semantic labels are disjoint from the label space of Cityscapes, thereby representing true out-of-distribution content. Objects are randomly scaled and positioned, with mammals more likely in the lower half and birds or airplanes in the upper half. FS Web is constructed similarly to FS Static, but uses automatically collected anomalous objects from the internet. The dataset is designed to be continuously updated. However, no more updates have been released since August 2022. To further evaluate the performance on real-world images, the author proposed \textbf{FS Lost \& Found} based on the original LostAndFound dataset \cite{pinggera2016lost}. This subset adds pixel-wise annotations for the roadside area and filter out objects appear in Cityscapes since they should not belongs to OoD instance \cite{cordts2016cityscapes}. To reduce redundancy, the authors subsampled highly repetitive sequences and excluded images without objects.

\textbf{SegmentMeIfYouCan} (SMIYC) was proposed by Chan et al. \cite{chan2021segmentmeifyoucan} in 2021, consists of RoadAnomaly21 and RoadObstacle21. RoadAnomaly21 provide an accessible validation set of 10 images and a private test set of 100 images, with image resolutions of either 1280 × 720 or 2048 × 1024. RoadObstacle21 offers an accessible validation set of 30 images and a private test set of 327 images, with a resolution of 1920 × 1080. \textbf{RoadAnomaly21} focuses on general anomaly segmentation in full street scenes and provides an evaluation set of 100 images with pixel-level annotations. This dataset extends RoadAnomaly dataset \cite{lis2019detecting} by removing low-quality images and labeling mistakes, introducing a \textit{void} class for ambiguous regions, and incorporating 68 newly collected images. Each image contains at least one anomalous object, and anomalies may appear anywhere in the scene across diverse environments. \textbf{RoadObstacle21} focuses on safety-critical scenarios in which obstacles appear directly on the road surface and pose immediate hazards to autonomous vehicles. To encompass diverse conditions, these images capture various road types, lighting conditions, and weather scenarios. Both RoadAnomaly21 and RoadObstacle21 use three annotation classes, namely anomaly or obstacle, non-anomalous or non-obstacle, and void.

% \textcolor{red}{A Dataset for Semantic Segmentation in the Presence of Unknowns}

% \begin{figure}[t]
%   \centering
%   %\fbox{\rule{0pt}{2in} \rule{0.9\linewidth}{0pt}}
%    \includegraphics[width=0.65\linewidth]{img/dataset_new.png}
%    \caption{Timeline of OoD segmentation dataset and their relationships.}
%    %\textcolor{red}{it would be better to avoid use light yellow and orange. Here your orange is too light.}
%    \label{fig:method}
% \end{figure}

\subsection{Evaluation Metrics}
Two commonly used metrics for evaluating OoD segmentation are the False Positive Rate at 95\% True Positive Rate (FPR95) and mean F1 score.

FPR95 measures the false positive rate on in-distribution pixels when the recall of OoD pixels reaches 95\%. It reflects how well a model avoids misclassifying in-distribution pixels as OoD under high-recall settings. It is defined as:

\begin{equation}
\text{FPR95} = \min_{\delta \in \mathbb{R}} \left( \frac{|\hat{P_{\mathrm{out}}}(\delta) \cap P_{\mathrm{in}}|}{P_{\mathrm{in}}} \ \Big| \ \text{recall}(\delta) \geq 0.95 \right),
\end{equation}
where $\hat{P_{\mathrm{out}}}$ denotes the predicted OoD pixels under threshold $\delta$.

To complement pixel-level evaluation, the mean F1-score provides a component-level perspective by comparing predicted and ground-truth OoD regions. It emphasizes whether entire OoD objects are successfully detected. The F1 score at threshold $\tau$ is computed as:
\begin{equation}
F1(\tau) = \frac{2 \cdot TP(\tau)}{2 \cdot TP(\tau) + FP(\tau) + FN(\tau)}.
\end{equation}
It is then averaged over multiple thresholds to obtain the final score.

\section{OoD Segmentation Methods}
In this section, we categorize existing OoD segmentation methods into four groups, based on the primary contribution of each work.
\label{sec:method}

\begin{figure}% 表示强制放置在当前位置
    \centering
    \includegraphics[width=1\linewidth]{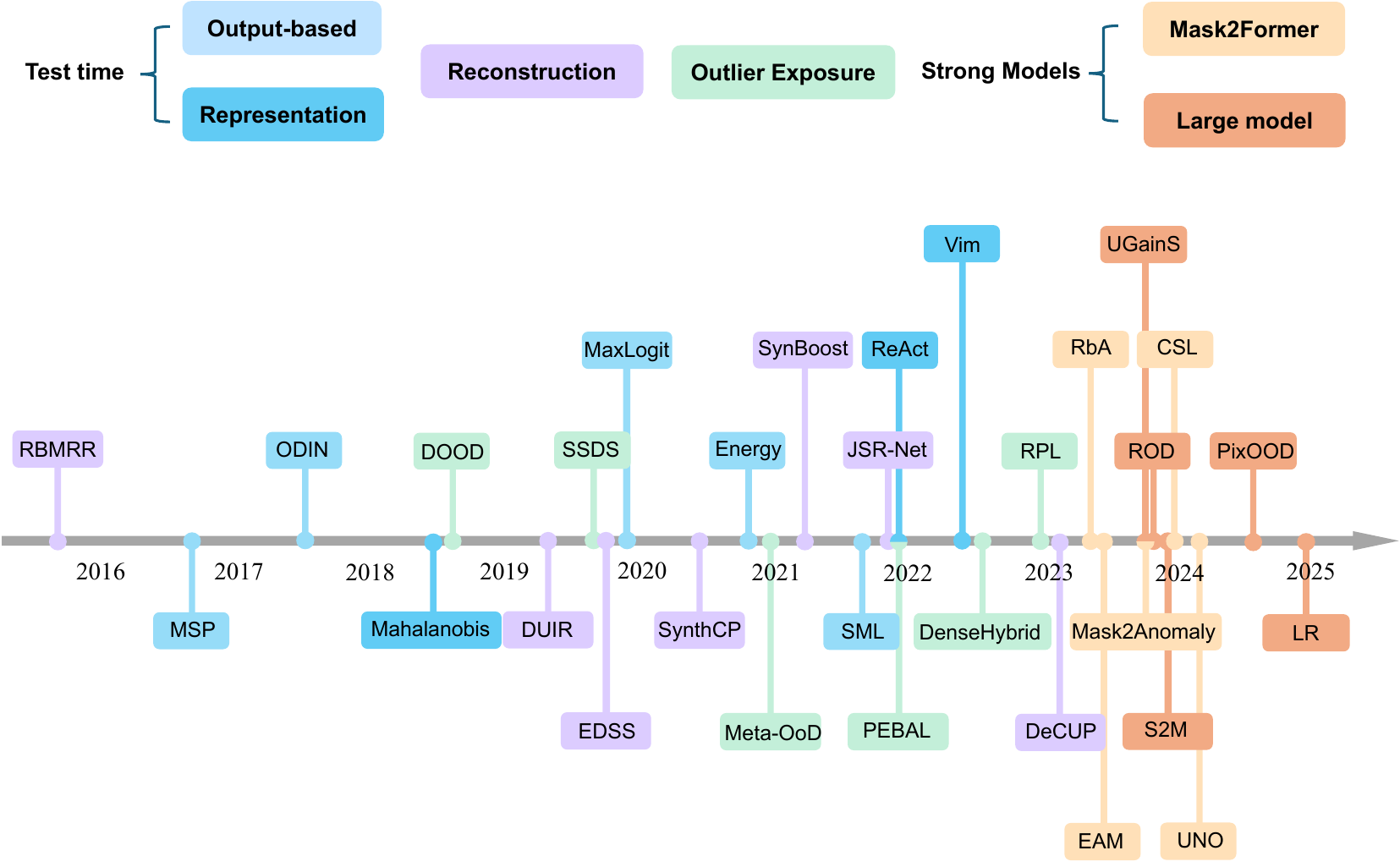}  % 替换为你图片的文件名
    \caption{Timeline of mentioned methods.}  % 图片标题
    \label{fig:timeline}  % 设置标签，用于引用
\end{figure}

\subsection{Test-time OoD Segmentation}
Some of these methods were originally developed for image-level OoD detection and have since been extended to address pixel-level segmentation tasks. More specifically, image segmentation is performed by assigning a class label to each individual pixel. The adaptation of these methods to OoD segmentation is achieved by computing anomaly score at the pixel level. This pixel-wise processing enables precise identification of out-of-distribution areas, which is particularly useful in autonomous driving.

\subsubsection{Output-based Estimation}

Output-based methods \cite{hendrycks2016baseline, hendrycks2019scaling, jung2021standardized, liang2017enhancing} were first introduced by Hendrycks and Gimpel along with the formulation of the OoD detection problem \cite{hendrycks2016baseline}. These methods rely on the model's output to compute an anomaly score, which is then used to identify OoD inputs.

Hendrycks et al. \cite{hendrycks2016baseline} used Maximum Softmax Probability (\textbf{MSP}) from the model’s output as the OoD score. The core idea is that ID inputs should have higher confidence scores, as the model should be able to classify them correctly. A model with OoD input should exhibits lower confidence, since it could not classify them correctly. This method successfully distinguishes ID and OoD inputs and serves as a baseline. However, it fails on large-scale datasets since probabilities will be diluted due to the presence of many similar. Consequently, the MSP score may be relatively low for some ID pixels, yet comparatively high for OoD regions that are dissimilar to any in-distribution class.
% \textbf{MSP} \textcolor{red}{16 Oct}

Maximum Logit (\textbf{MaxLogit}) was proposed to leverage logit values as OoD scores \cite{hendrycks2019scaling}. The motivation is that MSP suffers from competition among similar classes in the softmax distribution. Specifically, the logit values may be consistently high across similar ID classes, such as different types of cars, which results in low probabilities for each class. While OoD inputs may have smaller maximum logits compare to ID, the highest logits can still be significantly larger than all other classes. As a result, OoD inputs show a higher confidence score than certain ID inputs. MaxLogit avoids this competition and enables comparison between ID and OoD inputs. One challenge for this method is that the logit distributions differ across classes, which degrades its performance. Standardized Max Logits (\textbf{SML}) \cite{jung2021standardized} was proposed to address this issue by standardizing the maximum logits across classes, enabling more accurate comparison across class distributions.
% \textbf{MaxLogit}\textcolor{red}{19 Nov}
% \textbf{SML}\textcolor{red}{Jul 2021}

Another important advancement is \textbf{ODIN} \cite{liang2017enhancing}, which combines temperature scaling and small input perturbations to better separate in-distribution and out-of-distribution samples. Temperature scaling calibrates the softmax output by making the predicted probabilities softer, while input perturbations are designed to selectively increase the softmax score of ID inputs. Due to better calibration and sensitivity to the input, ODIN effectively enlarges the softmax score gap between ID and OoD samples. Despite their effectiveness, these output-based methods still struggle with the overconfidence issue. Neural networks tend to produce high confidence predictions even for out-of-distribution inputs. This overconfidence stems from the model’s logit outputs. As a result, OoD inputs may still receive high confidence scores, leading to incorrect identification. 

Liu et al. proposed the Energy score \cite{liu2020energy}, which leverages the logits of all classes. This score reflects how close an input is to the training distribution by measuring its total energy, and helps mitigate the impact of overconfidence. It provides a more robust confidence estimate and has been shown to outperform previous softmax-based scores. 
% \textcolor{red}{energy oct 2020}

\begin{table}[h]
\centering
\label{tab:ood_output_based}
\begin{tabular}{ll}
\toprule
\textbf{Method} & \textbf{Idea} \\
\midrule
\textbf{MSP} \cite{hendrycks2016baseline} 
& $
 s_{\mathrm{MSP}}(x) 
 = \max_{1 \le c \le K}
   \frac{e^{z_c(x)}}{\sum_{j=1}^K e^{z_j(x)}}
$ \\[8pt]
\textbf{MaxLogit} \cite{hendrycks2019scaling} 
& $
 s_{\mathrm{MaxLogit}}(x) 
 = \max_{1 \le c \le K} z_c(x)
$ \\[8pt]
\textbf{SML} \cite{jung2021standardized} 
& $
s_{\mathrm{SML}}(x) 
= \mathrm{Smooth}\Bigl(
    \mathrm{BoundarySuppress}\Bigl(
    % \mathrm{Standardize}\Bigl(
      \max_{1 \le c \le K}\,
        \frac{z_c(x) - \mu_c}{\sigma_c}
    % \Bigr)
    \Bigr)
  \Bigr)  
$ \\[8pt]
\textbf{ODIN} \cite{liang2017enhancing}
& $
 x^\prime 
= x \;-\; \epsilon \cdot \mathrm{sign}\!\Bigl(
  -\,\nabla_x 
  \log\!\Bigl(
    \max_{1 \leq c \leq K}
    \frac{\exp\!\bigl(z_c(x)/T\bigr)}
         {\sum_{j=1}^K \exp\!\bigl(z_j(x)/T\bigr)}
  \Bigr)
\Bigr)
$
\\[12pt]
\textbf{Energy Score} \cite{liu2020energy} 
& $
 s_{\mathrm{Energy}}(x) 
= -\,T \,\log \!\Biggl(
   \sum_{c=1}^K 
     \exp\!\Bigl(
       \frac{z_c(x)}{T}
     \Bigr)
\Biggr)
$ \\
\bottomrule
\end{tabular}
\caption{Summary of output-based OoD detection methods. 
\(\mathbf{z}(x)\in \mathbb{R}^K\) are the logits for input \(x\). For ODIN, we provide the formula for obtaining the preprocessed image.}
% Higher scores typically imply “more ID-like.”}
\end{table}

These output-based methods are easy to integrate into any model, making them widely adopted for OoD segmentation in autonomous driving. Moreover, they operate without requiring access to model parameters or training data, which is particularly valuable if the perception module of an automated driving systems is supplied by a third‑party provider. Nevertheless, these methodologies predominantly leverage model outputs, which may be insufficient for robustly distinguishing between ID and OoD samples.

\subsubsection{Representation-based Estimation} 
% \textcolor{red}{KNN embedding Representation-based Estimation.}
Instead of relying only on the model output, representation-based methods explore the intermediate features of neural networks to detect out-of-distribution samples. 
A representative method in this category is the \textbf{Mahalanobis Distance} approach proposed by \cite{lee2018simple}, whose effectiveness has been demonstrated on an autonomous-driving dataset in \cite{henriksson2023evaluation}. The idea is to model the feature distributions of each class using a multivariate Gaussian distribution. During testing, the Mahalanobis distance between a sample's feature and the nearest class distribution is calculated. Samples that are far from all class distributions are more likely to be OoD. This method leverages the information contained in the feature space. By analyzing how features deviate from known training distributions, it achieves more accurate detection compared to relying on final outputs alone. However, this approach requires access to the training set to compute feature deviations, limiting its applicability when the training data is private or inaccessible.
% \textcolor{red}{Mahalanobis Distance Jun 18}

Instead of extracting measurements directly from the feature space, other approaches focus on modifying the feature space to enhance output-based OoD scores. For example, Sun et al. proposed \textbf{ReAct} \cite{sun2021react}, which rectifies activations in the penultimate layer by imposing a threshold $c$. Specifically, the output of each neuron is clipped at $c$, taking the minimum value between the neuron's original activation and the threshold. An ideal $c$ should be enough to preserve the activation patterns for ID data while rectifying that of OoD data. By truncating overly high activations, the rectified activation can limit the effect of noise and make the overall activation pattern closer to a well-behaved case. This method effectively reduces the overconfidence issue commonly observed in neural networks. Furthermore, ReAct can be integrated with most commonly used OoD scores, making it easy to apply in OoD segmentation for autonomous driving. However, clipping also alters ID activations and can reduce ID accuracy. Even small drops matter in safety‑critical systems like autonomous driving. Therefore, ReAct requires extra validation before deployment.
% \textcolor{red}{ReAct 24 Nov 2021}

While representation-based methods leverage the rich information within the feature space, they do not guarantee optimal OoD detection for all inputs. Some OoD samples are easy to identify in the feature space, whereas others are easier in the logit space. To effectively combine these two perspectives, Wang et al. introduced \textbf{ViM} \cite{wang2022vim}. It generates an additional virtual logit representing a synthetic OoD class derived from feature residuals. Then, this virtual logit is scaled to match the magnitude of the original logits, ensuring compatibility. During inference, the probability corresponding to this virtual class serves as an effective OoD score. These methods utilize the rich, discriminative information embedded in deep features and can be deployed without re-training the backbone network. However, they typically require access to training statistics, involve hyperparameters such as threshold $c$, which complicates deployment.

The experimental results of test-time OoD segmentation methods are presented in the first part of Table \ref{tab:comparison}. These methods introduce various OoD scoring functions that have been widely adopted and now serve as the foundation for most subsequent OoD segmentation techniques.

% \textcolor{red}{PixOOD: Pixel-Level Out-of-Distribution Detection}

\subsection{Outlier Exposure for Supervised Training}

% \begin{figure}% 表示强制放置在当前位置
%     \centering
%     \includegraphics[width=1\linewidth]{img/oe.png}  % 替换为你图片的文件名
%     \caption{Motivation for Outlier Exposure in Supervised Training. Outlier exposure introduces proxy OoD samples by pasting objects from an outlier dataset into ID images. The resulting images can then be used for supervised training to improve OoD segmentation performance. \textcolor{red}{consider half column for this picture}}  % 图片标题
%     \label{fig:oe}  % 设置标签，用于引用
% \end{figure}

\begin{wrapfigure}{r}{0.5\linewidth} 
% \vspace{-8mm}

% 'r' = right side; adjust width as needed
  \centering
  \includegraphics[width=0.9\linewidth]{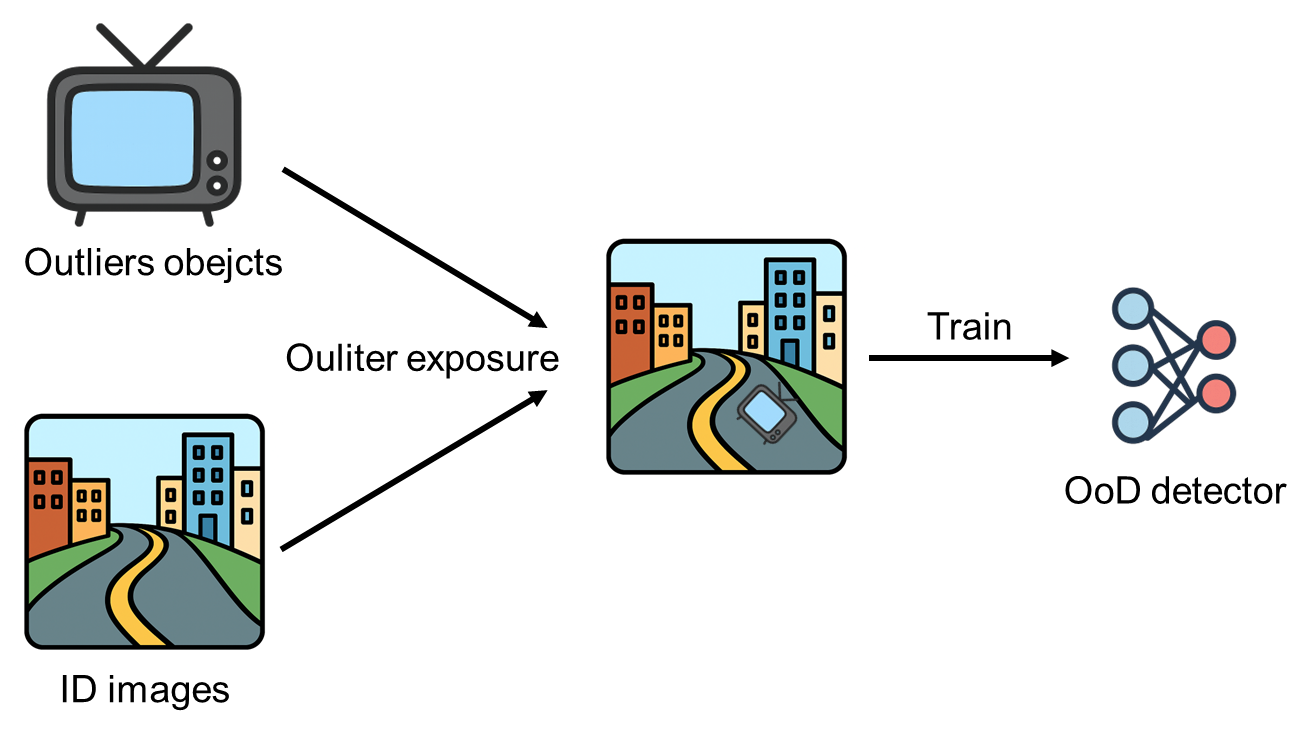}
  \caption{Motivation for Outlier Exposure in Supervised Training. Outlier exposure introduces proxy OoD samples by pasting objects from an outlier dataset into ID images. The resulting images can then be used for supervised training to improve OoD segmentation performance.}
  \label{fig:oe}

% \vspace{-7mm}
\end{wrapfigure}

With the established assumptions, models are expected to remain unexposed to OoD data during training. Nevertheless, training is the intuitive approach for amplifying the distinction between ID and OoD data. To make OoD detection training feasible, Hendrycks et al. proposed Outlier Exposure (OE) \cite{hendrycks2018deep}, which incorporates an auxiliary dataset of outliers that does not overlap with real OoD data that may occur during testing. Their loss function is designed to preserve the original task objectives while simultaneously encouraging the model to assign low confidence scores to outlier samples. Outlier Exposure significantly improves OoD detection performance on various tasks. The impressive performance of image‑level outlier exposure suggests that a similar strategy may also benefit pixel‑wise OoD segmentation in autonomous‑driving applications. Figure \ref{fig:oe} illustrates the OE process for generating a dataset that can be used to improve OoD segmentation.

Bevandic et al. treat pixel‑level OoD detection as a binary classification task \cite{bevandic2018discriminative}. We denote this approach as \textbf{DOOD}. To train the discriminative model, they incorporate outlier exposure into semantic segmentation datasets. They select a road-driving dataset such as Cityscapes as ID data, where every pixel in these images is labeled as ID. Objects from ImageNet are then pasted on top of ID images as OoD pixels. This process generates a comprehensive synthetic dataset in which ID and OoD regions coexist within the same frame, making supervised training feasible. Using this synthetic dataset, they train a dedicated OoD discriminative model that identifies whether each pixel belongs to ID or OoD categories. Experimental results show that this supervised training enables the detector to significantly outperform softmax-based baselines. However, the dedicated OoD segmentation model focuses solely on detecting OoD pixels and ignores the segmentation task.
% \textcolor{red}{DOOD  Aug 2018}

Following their earlier discriminative detector, Bevandic et al. introduce a unified architecture that outputs both semantic segmentation results and a binary OoD map in a single forward pass \cite{bevandic2019simultaneous}, we denote this approach as \textbf{SSDS}. They append a lightweight outlier detection head to a standard segmentation backbone. This extra head  is trained on an outlier dataset, allowing the model to identify pixels that belong to OoD regions. Although the reported results are promising, multi-task training can decrease segmentation accuracy. Moreover, the abnormally high presence of OoD pixels in OE datasets risks introducing an OoD‑frequency bias. Such bias can hinder deployment in autonomous driving scenarios.
% \textcolor{red}{ssds  Aug 2019]}

Similarly, Chan et al. introduce an entropy maximization loss applied to synthetic datasets generated through outlier exposure in \textbf{Meta-OoD} \cite{chan2021entropy}. They incorporate a meta classifier as a post-processing step to filter out false positive OoD samples. The meta classifier uses handcrafted metrics computed per connected component, based on pixel-wise uncertainty measures derived from the softmax probabilities, to identify incorrect OoD segmentation. Experimental results demonstrate that this two-stage approach effectively reduces OoD detection errors. Additionally, the authors explore the impact of re-training on the original semantic segmentation task. Their findings indicate that re-training leads to a measurable but marginal decrease in segmentation accuracy. Besides, meta-OoD tends to produce fragmented anomaly masks with many false positive pixels filtered out. It relies on re-training the entire network for outlier exposure, which is inefficient and risks biasing predictions towards outliers.
% \textcolor{red}{METAOOD  Dec 2020, only detection,have acc, suitable]}

To decouple anomaly detection from uncertainty and reduce computational cost, Tian et al. propose pixel‑wise energy‑biased abstention learning (\textbf{PEBAL}) \cite{tian2022pixel}. PEBAL adds a pixel‑level anomaly class that the model can choose whenever a pixel does not match any inlier category. It jointly optimizes a pixel‑wise abstention‑learning head and an energy‑based model. Energy‑based model is trained to assign high energy to anomaly pixels drawn from an outlier dataset, which decomposes out-of-distribution pixels with high uncertainty. Pixels with high energy get a larger penalty, so pixel‑wise abstention‑learning is more inclined to categorize them as anomalies. In addition, a smoothness loss encourages neighboring anomaly pixels to be consistent, and a sparsity loss reduces false positive detections. Because only the last block of the backbone is fine‑tuned, the approach is lightweight for the automated driving systems. Experiments on various datasets show that PEBAL improves OoD segmentation with fewer false positives than Meta-OoD. Nevertheless, retraining process sacrifices the ID segmentation accuracy,  which is crucial in autonomous driving. 
% \textcolor{red}{PEBAL  Nov 2021]}

To better leverage both discriminative approaches and density estimation methods, \textbf{DenseHybrid} \cite{grcic2022densehybrid} incorporates these two perspectives into a hybrid anomaly detector. The model is trained on the OE dataset with loss a function composed of three objectives: classification loss $L_{cls}$, unnormalized data likelihood loss $L_x$, and dataset posterior loss $L_d$. They proposed maximizing the likelihood of in-distribution pixels and minimize the likelihood of out-of-distribution pixels. The dataset posterior prediction $P(d_{out}|\mathrm{x})$ represents the probability that a pixel belongs to the out-of-distribution. The authors define the hybrid anomaly detector $s(\mathrm{x})$ based on the out-of-distribution dataset posterior prediction $P(d_{out}|\mathrm{x})$ and the likelihood prediction $p(\mathrm{x})$ as:
$$s(\mathrm{x}) \cong \ln P(d_{out} | \mathbf{x}) - \ln \hat{p}(\mathbf{x}).
$$
The anomaly score map is then fused with semantic segmentation output based on a threshold. DenseHybrid achieves better out-of-distribution segmentation with a smaller sacrifice in in-distribution segmentation accuracy. Nevertheless, the method still introduces a slight degradation in ID segmentation performance, which may affect its applicability in autonomous driving.
% \textcolor{red}{DenseHybrid  6 Jul 2022]}

A critical issue is that OoD detectors may incorrectly assign high OoD scores to in-distribution pixels under covariate shifts in the environment, such as when the scene changes from urban to rural environments. Liu et al. propose Residual Pattern Learning (\textbf{RPL}) \cite{liu2023residual} to fully the re-training impact on in-distribution segmentation and to generalize across various contexts. They introduce an external module, RPL, attached to a frozen closed-set segmentation network as the OoD detector. The vanilla segmentation network is represented as:
$$
\tilde{\mathrm{y}} = f_{\phi_{\text{seg}}} \left( f_{\phi_{\text{aspp}}} \left( f_{\phi_{\text{fcn}}} (\mathbf{x}) \right) \right).
$$
RPL takes intermediate features as input to learn the residual patterns of anomalies. Its output is fused with the segmentation features to induce high uncertainty for out-of-distribution regions. The modified network is represented as: 
$$
\hat{\mathrm{y}} = f_{\phi_{\text{seg}}} \left( f_{\phi_{\text{aspp}}} \left( f_{\phi_{\text{fcn}}} (\mathbf{x}) \right) + f_{\theta_{\text{rpl}}} \left( f_{\phi_{\text{fcn}}} (\mathbf{x}) \right) \right).
$$
The in-distribution segmentation accuracy is guaranteed by freezing the segmentation model. To enhance the models' understanding of the relationship between context and out-of-distribution objects, the authors introduce context-robust contrastive learning (CoroCL), whose objective is to make embeddings of the same distribution more similar and embeddings of different distributions more distinct. 
At inference time, the in-distribution segmentation result is obtained from the original network branch $\tilde{\mathrm{y}}$, while the out-of-distribution segmentation result $\hat{\mathrm{y}}$ is computed with RPL module. 
RPL \cite{liu2023residual} demonstrates strong out-of-distribution segmentation performance without sacrificing in-distribution segmentation accuracy. However, the two output maps may sometimes be inconsistent. Specifically, the detected out-of-distribution region on the anomaly map may not correspond to any object in the segmentation map, which can potentially confuse decision-making modules in automated driving systems.
% \textcolor{red}{RPL Nov 2022]}

\begin{table}[ht]
\centering
\footnotesize
\setlength{\tabcolsep}{5pt}
\begin{tabular}{@{}lcccc@{}}
\toprule
\textbf{Method} & \textbf{Year} & \textbf{ID influence} & \textbf{Retrain backbone} & \textbf{Key idea}\\
\midrule
DOOD\,\cite{bevandic2018discriminative}          & 2018 & - & \hmark    & Discriminative binary classifier on synthetic ID+OoD mix \\[2pt]
SSDS\,\cite{bevandic2019simultaneous}            & 2019 & \textcolor{dropYes}{Yes} & \cmark\ & Extra detection head joint-trained with seg. network \\[2pt]
Meta-OoD\,\cite{chan2021entropy}                 & 2021 & \textcolor{dropYes}{Yes}           & \cmark\ & Entropy-max + meta-classifier to filter FP \\[2pt]
PEBAL\,\cite{tian2022pixel}                      & 2022 & \textcolor{dropYes}{Yes}           & \hmark & Energy-biased abstention; sparse + smooth losses \\[2pt]
DenseHybrid\,\cite{grcic2022densehybrid}         & 2022 & \textcolor{dropYes}{Yes}            & \cmark\ & Likelihood + posterior hybrid anomaly score \\[2pt]
RPL\,\cite{liu2023residual}                      & 2023 & \textcolor{dropNo}{No} & \xmark\ & External residual pattern head + CoroCL \\ 
\bottomrule
\end{tabular}
\caption{Summary of OE-based supervised OoD segmentation methods. The \textbf{ID influence} column indicates whether training affects ID segmentation, while in the \textbf{Retrain backbone} column \cmark, \hmark, and \xmark\ denote fine-tuning most backbone layers, updating only the last block, and keeping the backbone frozen, respectively.}
\label{tab:oe_methods}
\end{table}

Training on a synthetic dataset from OE significantly improves the OoD segmentation performance. However, it remains questionable whether the outlier datasets employed during outlier exposure can accurately simulate the OoD objects that occur in real-world driving scenarios. Moreover, the sacrifice of in-distribution segmentation accuracy may not be acceptable in perception module. Even if the model produces two separate outputs, one for in-distribution segmentation and one for OoD segmentation, the conflict between these outputs may confuse the decision-making module in automated driving systems.

Table \ref{tab:oe_methods} provides a detailed comparison of the OE-based supervised OoD segmentation methods discussed above. The experimental results of supervised training based on outlier exposure are presented in the second part of Table \ref{tab:comparison}. While these methods significantly improve performance, leveraging outlier dataset as a proxy for real-world OoD data introduces a strong assumption of OoD samples. In practice, real-world OoD data is unpredictable. The introduction of OE dataset may limit the generalizability in real-world scenarios.

\subsection{Reconstruction-based OoD Segmentation }

\begin{figure}[ht]% 表示强制放置在当前位置
    \centering
    \includegraphics[width=1\linewidth]{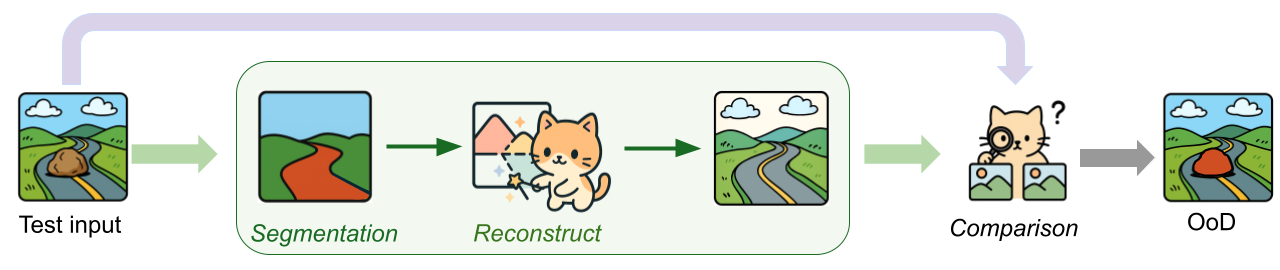}  % 替换为你图片的文件名
    \caption{Motivation for reconstruction-based OoD segmentation. OoD objects may be ignored by the segmentor and thus omitted in the reconstruction. Comparing the original image and the reconstruction enables OoD segmentation.}  % 图片标题
    \label{fig:recon}  % 设置标签，用于引用
\end{figure}

A well-trained model should be able to reconstruct the original image. The difference between the reconstructed image and the original image may indicate potential regions of road obstacles. Based on this idea, reconstruction-based methods have emerged as another mainstream approach for OoD segmentation in autonomous driving scenarios. These methods segment anomalies by comparing the input image to a reconstructed version that represents the expected in-distribution appearance, as shown in Fig. \ref{fig:recon}. The differences between these two images are more likely to correspond to anomalous regions.

Creusot et al. used a trained Restricted Boltzmann Machine to reconstruct the image in \cite{7225680}. We denote this work as \textbf{RBMRR}. The reconstruction result may miss small obstacles in the background. For example, the model may only reconstruct the road surface even when the input image contains an obstacle on the road. The difference between the reconstructed and original images is used to perform anomaly segmentation. Here the input to the reconstruction consists of normalized patches. The reconstruction vector $x'$ is computed as
$$x'_i = \mathrm{Sigmoid}(x_i \cdot W + b_{\mathrm{hid}}) \cdot W^{T} + b_{\mathrm{vis}},
$$where $W$, $b_{hid}$ and $b_{vis}$ are parameters of Restricted Boltzmann Machine.
This work verifies the feasibility of reconstruction-based OoD segmentation and demonstrates its effectiveness on both texture-based synthetic datasets and real-world images in highway scenarios. However, the low quality of the reconstructed images limits the detection performance. The patch-based reconstruction also inherently limits the model’s ability to adapt, resulting in suboptimal performance when detecting larger objects.
% \textcolor{red}{RBMRR August 2015], no train need with provided beckbone and re-synthesis module.} 

With the development of semantic segmentation, networks can generate precise semantic maps for input images. However, obstacles on the road may be ignored by segmentation networks, which inspires Lis et al. \cite{lis2019detecting} to reconstruct the original image from the predicted semantic map.This method is referred to as \textbf{DUIR} in the literature. They proposed resynthesizing the original image based on the semantic prediction, where segmentation errors lead to significant differences between the resynthesized and original images. To achieve this, they employ \textit{pix2pixHD} \cite{wang2018high}, a conditional GAN \cite{goodfellow2014generativeadversarialnetworks} trained on in-distribution dataset, to generate the image conditioned on semantic map. In their discrepancy network, features are extracted from both the original and resynthesized images using a pre-trained VGG network, while a separate CNN processes the one-hot representation of the predicted labels. At each level of the feature pyramid, these features are concatenated and passed through 1$\times$1 convolutions. Additionally, pointwise correlations between features of the original and resynthesized images are computed and incorporated. All the fused features and correlations are then fed into an up-convolutional decoder to produce the final discrepancy score.
% \textcolor{red}{DUIR Apr 2019] need train} 

Similarly, Xia et al, proposed a unified framework \textbf{SynthCP} that addresses both failure and anomaly detection in semantic segmentation through a synthesize-and-compare strategy \cite{xia2020synthesize}. A semantic-to-image conditional GAN \cite{park2019semantic} is trained on in-distribution data to reconstruct the input image from the predicted segmentation result. Both the original and synthesized images are then fed into the comparison module, which computes the cosine distance between the corresponding feature vectors extracted from the last layer of the segmentation model at each pixel. To overcome the inherent misclassifications in the segmentation model, the authors refine the comparison result $\hat{c}_n^{(i)}$ using the maximum softmax probability: 
$$
\hat{c}_n^{(i)} \leftarrow \hat{c}_n^{(i)} \cdot \mathbb{I}\{ p^{(i)} \leq t \} + (1 - p^{(i)}) \cdot \mathbb{I}\{ p^{(i)} > t \},
$$ where $p^{(i)}$ is the maximum softmax probability at the $i$-th pixel, $t \in [0,1]$ is a threshold and $\mathbb{I}$ denotes the indicator function. This post-processing filters out pixels where the segmentation predictions have high confidence. This method relies on numerical discrepancies between the original and reconstructed images. However, it lacks a comprehensive understanding of the semantic regions. 
% \textcolor{red}{18 Mar 2020, need train} 

Although the difference between original and reconstructed image often indicates OoD regions, pixel-level comparison may not be an effective way to measure semantic discrepancy. For example, two images with large differences in brightness may yield high pixel-wise discrepancy. However, they are very similar to the human eye at the semantic level. Similar issues arise in images generated by GANs, where such variations can lead to false positives. Haldimann et al. identified and address this limitation by proposing a CNN-based dissimilarity detector to estimate the semantic inconsistency between the original and reconstructed images \cite{haldimann2019not}. We denote this work as \textbf{EDSS}. They first resynthesize an RGB image from the predicted semantic map using \textit{pix2pix} \cite{isola2017image}, a conditional GAN trained on in-distribution data. To measure discrepancy at the semantic level, they train a dissimilarity detector that compares original and reconstructed images. It uses VGG-based feature extraction and is optimized with a triplet loss that assigns higher dissimilarity to mismatched patches and lower values to matched pairs. The loss is defined as
$$
\mathcal{L}(D) = \lambda_D \, \mathbb{E}_{t_i}[\log D(p^+_i)] + \mathbb{E}_{t_i}[\log(1 - D(p^-_i))],
$$ where $D$ is dissimilarity detector, $p^+_i$ and $p^-_i$ denote the positive and negative patch pairs respectively.
A higher anomaly score in dissimilarity map means that a pixel is more likely to belong to an out-of-distribution object. This approach leverages semantic information beyond simply comparing pixel-level discrepancies. However, it remains limited to comparing the original image with the resynthesized one.
% \textcolor{red}{fixed GAN, pix2pix, compare generate and original,we name it Dissim. 2019/9/2, need train}

To better leverage the semantic information for out-of-distribution segmentation, Vojir et al. proposed \textbf{JSR-Net} \cite{vojir2021road}, which includes a segmentation coupling module to fuse information from the segmentation map and the reconstruction error. Instead of using GAN, the authors employ a lightweight bottleneck decoder to reconstruct road areas. The similarity between the original image and reconstructed image is measured by SSIM \cite{wang2003multiscale}. The reconstruction module is trained to maximize the similarity between reconstructed and original images in the road area and minimize it on anomalies. 
The segmentation coupling module first concatenates the semantic segmentation logits with the reconstruction error. It then processes this input with two convolution blocks to predict a binary mask that distinguishes road regions from anomalies. The segmentation coupling module is optimized with binary cross entropy loss $\mathcal{L}_{xent}$.
The final loss function is defined as 
$$
\mathcal{L} = \mathcal{L}_{\text{xent}} + 0.5\mathcal{L}_{R}, 
$$ where $\mathcal{L}_{R}$ is reconstruction loss. This method reduces compute cost and improves out-of-distribution segmentation performance. However, it still suffers from false positives on thin structures and is sensitive to image quality degradation. The authors extend this work by introducing several subtle components in \textbf{DaCUP} \cite{vojivr2023image}. To better utilize data and capture appearances of road surfaces, the authors inject an embedding network optimized with triplet loss. In addition, it designs a distance-based scoring mechanism in the embedding space and an inpainting module to reduce false positive rate. These components improve out-of-distribution segmentation performance. However, the pipeline is complex and contains multiple modules.
% \textcolor{red}{JSR net Oct 2021, need train}
% \textcolor{red}{DaCUP Jan 23, need train}

Reconstruction-based methods may fail when the segmentation output is noisy and fragmented. To better leverage both reconstruction-based methods and diverse OoD scores, Biase et al. propose \textbf{SynBoost} \cite{di2021pixel}, a novel framework that combines these two components to produce more robust OoD segmentation. It integrates reconstruction methods with uncertainty measures: softmax entropy \cite{gal2016uncertainty, lakshminarayanan2017simple}, softmax difference \cite{rottmann2020prediction}, and perceptual differences \cite{johnson2016perceptual, dosovitskiy2016generating}. It uses CC-FPSE \cite{liu2019learning}, a conditional GAN, to reconstruct the original image from the predicted semantic segmentation map. The softmax entropy and softmax difference are computed from output of segmentation network. To avoid low-level texture differences in the synthesized image, such as differences in cars colors, the authors compute perceptual differences between corresponding pixels $x$ and $r$ as follows:
$$
\ V(x, r) = \sum_{i=1}^{N} \frac{1}{M_i} \| F^{(i)}(x) - F^{(i)}(r) \|_1,
$$
where $F^{(i)}$ denotes the $i$-th layer with $M_i$ elements of the VGG network, and $N$ is the number of layers \cite{tian2022pixel}. 
These inputs are then passed to a dissimilarity model that predict a mask of out-of-distribution objects. The model uses VGG to extract features from the original and synthesized images, and a CNN to process semantic predictions and uncertainty maps. Then, features from the input image, the reconstruction, and the segmentation output are concatenated at each level and passed through a $1 \times 1$ convolution to compare them. After that, resulting feature map will be fused with uncertainty map using point-wise correlation. Finally, a decoder leverages fused features across levels along with the semantic map to produce the final OoD segmentation prediction.
% \textcolor{red}{SynBoost 9 Mar 2021, need train}

Table \ref{tab:reconstruction_methods} provides a concise comparison of the reconstruction-based OoD segmentation pipelines discussed in this section. The experimental results of reconstruction-based methods are presented in the third part of Table \ref{tab:comparison}. These methods were popular for a period of time but have seen a decline in interest within the community.

% \begin{table}[htbp]
% \centering
% \caption{Comparison of reconstruction-based OoD segmentation methods. \textcolor{red}{The pipeline here is a little bit arguly, polish it. icons are beautiful I think}}
% \resizebox{\textwidth}{!}{
% \begin{tabular}{lcccc}
% \toprule
% \textbf{Method} & \textbf{Generator Model} & \textbf{Comparison Method} & \textbf{Feature Extraction} & \textbf{Semantic Info in Comparison} \\
% \midrule
% RBMRR \cite{7225680}   & Restricted Boltzmann Machine & Pixel-wise difference & N & N \\
% DUIR \cite{lis2019detecting}   & pix2pixHD (cGAN)  & CNN-based discrepancy network & Y & Y \\
% SynthCP~\cite{xia2020synthesize}  & SPADE (cGAN)      & Cosine distance + threshold & Y & Y (partially) \\
% EDSS~\cite{haldimann2019not}     & pix2pix (cGAN)    & CNN-based dissimilarity detector & Y & Y \\
% JSR-Net~\cite{vojir2021road}  & Bottleneck Decoder           & SSIM + segmentation coupling module & N & Y (partially) \\
% DaCUP~\cite{vojivr2023image}    & JSR-Net & Embedding-space distance & Y & Y \\
% SynBoost~\cite{rai2023unmasking} & CC-FPSE (cGAN)    & VGG perceptual difference & Y & Y \\
% \bottomrule
% \end{tabular}}
% \label{tab:reconstruction_methods}
% \end{table}

\begin{table}[htbp]
\centering
\resizebox{\textwidth}{!}{
\begin{tabular}{lcccccc}
\toprule
\textbf{Method} & \textbf{Generator Model} & \textbf{Comparison Method} &
\textbf{Feature Extraction} & \textbf{Gen.\ Training} &
\textbf{Semantic Info in Comparison} \\  
\midrule
RBMRR~\cite{7225680}         & Restricted Boltzmann Machine & Pixel-wise diff. &
N & P& N \\
DUIR~\cite{lis2019detecting} & pix2pixHD (cGAN) % train on cityscape, 不算完全general 的generator
& CNN discrepancy net &
Y & P & Y \\
SynthCP~\cite{xia2020synthesize} & SPADE (cGAN)              & Cosine + threshold &
Y & P & N \\          % 只在重建阶段用语义 → N
EDSS~\cite{haldimann2019not} & pix2pix (cGAN) % train on cityscape, 不算完全general 的generator
& Dissimilarity detector      &
Y & P & N \\
JSR-Net~\cite{vojir2021road} & Bottleneck decoder           & SSIM + coupling       &
N & J & Y \\
DaCUP~\cite{vojivr2023image} & Bottleneck decoder           & SSIM + Embedding-space dist.  &
Y & J & Y \\
SynBoost~\cite{di2021pixel} & CC-FPSE (cGAN)           & Decoder blocks   &
% synthesis module train on ID
Y & O & Y \\
\bottomrule
\end{tabular}}
\caption{Comparison of reconstruction-based OoD-segmentation pipelines.  
\textbf{Feature Extraction} indicates whether the comparison module relies on an feature extractor.
\textbf{Gen.\ Training} summarises how the generator is handled: \textbf{O}—off-the-shelf weights from other works; \textbf{P}—first pre-trained on ID data and then frozen for detection; \textbf{J}—jointly optimised together with the detection head. 
\textbf{Semantic Info in Comparison} specifies whether semantic cues are  incorporated during comparison.}
\label{tab:reconstruction_methods}

\end{table}

\subsection{Powerful Models for OoD Segmentation}

\begin{figure}% 表示强制放置在当前位置
    \centering
    \includegraphics[width=1\linewidth]{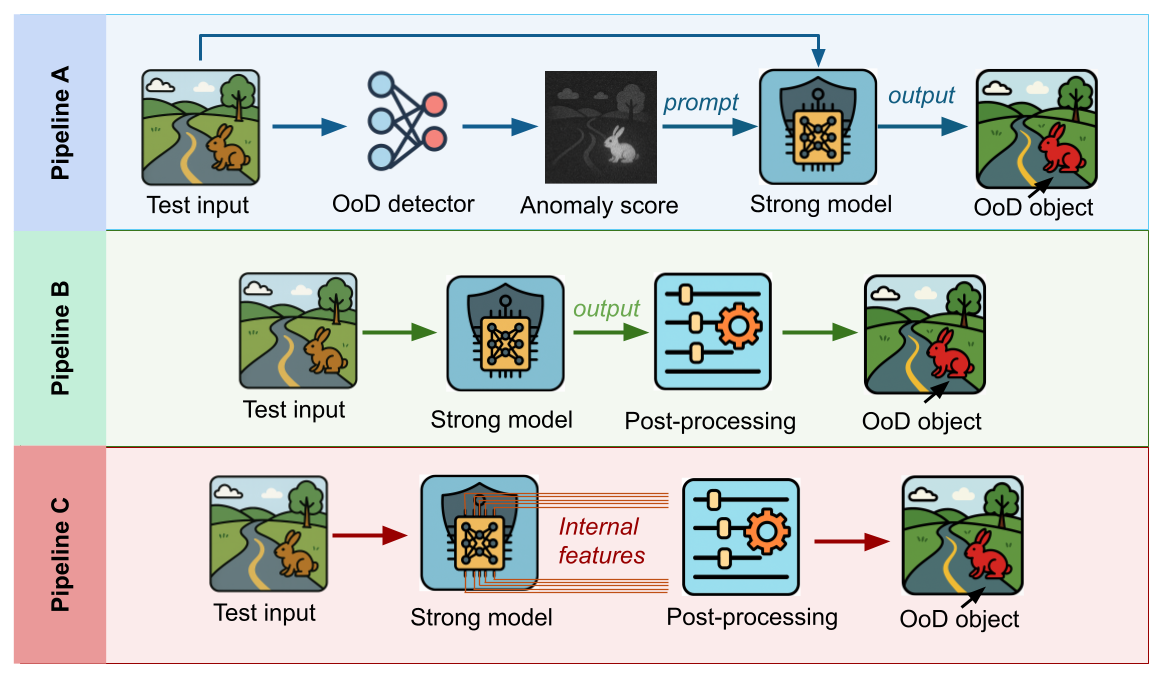}  % 替换为你图片的文件名
    \caption{Typical pipelines for OoD segmentation using powerful pretrained models. Pipeline A uses an external OoD detector, Pipeline B uses the final model output, and Pipeline C uses intermediate features}
    \label{fig:strong}  % 设置标签，用于引用
\end{figure}

A significant challenge shared by almost all of the previously discussed methods is that their outputs are often fragmented and cannot be interpreted at the semantic level. However, autonomous driving is more concerned with semantic-level understanding than whether some pixels in this object belong to in-distribution while other pixels belong to out-of-distribution. The development of the semantic segmentation community has brought some powerful models which can give precise masks for images. It is a trend to leverage these powerful segmentation models to segment OoD objects precisely in different ways. Moreover, CLIP \cite{radford2021learning} unifies the feature spaces of natural language and images, which makes it possible to segment out-of-distribution objects with the help of textual information. 
% In this section, we discuss methods that leverage either strong segmentation models or vision-language models via different pipelines, as shown in Fig. \ref{fig:strong}.
In this section, we review methods that leverage powerful segmentation backbones or vision–language models through different pipelines, as illustrated in Fig. \ref{fig:strong}.

\subsubsection{Mask2Former-Based}
% \textcolor{red}{Maskomaly: Zero-shot mask anomaly segmentation}
Mask2Former \cite{cheng2022masked} is a universal image segmentation framework designed for semantic, instance, and panoptic segmentation using a masked attention transformer decoder. \textbf{Mask2Anomaly} \cite{rai2023unmasking} builds upon Mask2Former \cite{cheng2022masked} and reformulates anomaly segmentation as a mask classification task instead of per-pixel classification. To improve OoD segmentation at the object level, the authors extend Mask2Former \cite{cheng2022masked} with three key contributions, resulting in the Mask2Anomaly framework. First, it introduces a global mask attention mechanism that helps the model focus on both foreground and background. During training, it uses contrastive learning with synthetic dataset from OE to encourage the model to maximize the margin between the anomaly scores of ID and OoD classes. Finally, a refinement module is applied during inference to filter out false positive regions based on the panoptic segmentation result. Mask2Anomaly significantly reduces the false positive rate and improves performance across datasets. The complex module design built upon Mask2Former \cite{cheng2022masked} makes Mask2Anomaly incompatible with plug-and-play deployment with advanced segmentation models. 
% \textcolor{red}{8 Sep 2023 }

\textbf{RbA} \cite{nayal2023rba} was proposed by Nayal et al. to address the limitations of pixel-wise OoD segmentation. Instead of modifying Mask2Former \cite{cheng2022masked}, RbA leverages the output of Mask2Former and introduces a scoring function to identify masks rejected by all known classes. Specifically, the method aggregates the class probabilities of regions and mask outputs to estimate the classification probability of each pixel. To measure the masks rejected by all known classes, the authors assume class label $K+1$ represents the region of out-of-distribution objects. This outlier probability is defined as:
$$
p(\mathbf{y} = K + 1 \mid \mathbf{x}) =1 - \sum_{k=1}^{K} p(\mathbf{y} = k \mid \mathbf{x}).
$$
A higher outlier probability indicates that pixels belonging to out-of-distribution objects have low predicted probabilities for all known classes. RbA leverages pre-trained Mask2Former and shows strong performance without training on outlier data. A lightweight fine-tuning variant which only optimizes a small portion of the model parameters surpasses state-of-the-art performance without degrading the closed-set performance. However, the $K+1$-th logit in Mask2Former is originally designed to represent the 'no object' class, which may not accurately correspond to anomalous regions and conflate uncertainty with outlier recognition.
% RbA \cite{nayal2023rba} \textcolor{red}{23 March}. 

\textbf{EAM}, similar to RbA \cite{grcic2023advantages}, also recognizes the potential of leveraging a mask-level recognition model in OoD segmentation. It proposes the EAM outlier detector incorporates negative supervision to achieve competitive results. EAM estimates uncertainty by aggregating low-confidence predictions across multiple masks.
% \textcolor{red}{23 Apr} 

To disentangle the uncertainty prediction and negative object classification, Delic et al proposes \textbf{UNO} \cite{delic2024outlier} which integrates an uncertainty score and a negative score. For out-of-distribution segmentation scenarios, it extends the Mask2Former to $K+2$ classes and fine-tunes on outlier dataset where $K+1$-th logit represents negative objects and $K+2$-th logit represent the no-object class. Specifically, the uncertainty score $S_{Unc}(\textbf{z})$ is defined as the negative maximum softmax probability over the known $K$ classes:

$$
\mathbf{s}_{\text{Unc}}(\mathbf{z}) := - \max_{k = 1 \ldots K} P(Y = k \mid \mathbf{z})
$$
The negative objectness score $S_{NO}$ is defined as:
$$
\mathbf{s}_{\text{NO}}(\mathbf{z}) := P(Y = K+1 \mid \mathbf{z})
$$
The UNO score is defined as the sum of these two scores. UNO decouples negative prediction from uncertainty prediction and attains strong
performance on both pixel level and image level benchmarks. However, extra class may introduce bias into uncertainty estimates.
% \textcolor{red}{24 Feb}

Zhang et al. propose Class-Agnostic Structure-Constrained Learning
(\textbf{CSL}) \cite{zhang2024csl}, a plug-in framework to embed structural constraints. It offers two integration schemes, one involves distilling knowledge from a teacher network end-to-end to the CSL framework, and the other applies those constraints during inference by fusing region proposals with pixel-level predictions. Additionally, CSL introduces a mask split preprocessing to break semantic masks into isolated components, thereby attenuate class bias. CSL improves performance on OoD segmentation, zero-shot semantic segmentation and domain adaptation. However, it significantly increases architectural complexity compared to more straightforward methods.
% \textcolor{red}{23 Dec}

These methods leverage mask-level predictions from Mask2Former to produce precise masks for out-of-distribution objects. However, Mask2Former lacks strong generalization to novel classes and struggles to accurately segment objects it has never encountered during training. This limitation affects all Mask2Former-based methods employed in automated driving systems, since novel objects can appear at any time.

\subsubsection{Large model}
Large models such as the Segment Anything Model (SAM) \cite{kirillov2023segment}, CLIP \cite{radford2021learning}, and the DINO series \cite{caron2021emerging, oquab2024dinov2learningrobustvisual, liu2024grounding} provide powerful capabilities for open-world segmentation. SAM was pretrained on over 11 million images and 1 billion masks, achieves zero-shot segmentation with prompt input. CLIP aligns visual and textual embeddings, enabling models to reason about novel categories via natural language descriptions. DINO series serve as strong foundation models for feature extraction. These models present promising opportunities for advancing out-of-distribution segmentation. 

\textbf{UGainS} utilizes SAM with point prompts \cite{nekrasov2023ugains}. It first computes an anomaly score map using RbA \cite{nayal2023rba}. Point prompts are then sampled via a farthest point sampling strategy. SAM receives the original image and these point prompts to generate masks for out-of-distribution objects. Although SAM can produce highly precise masks when prompted correctly, its performance is limited by the quality of the sampled points. If the anomaly score map is noisy, the resulting prompts may be suboptimal and degrade overall performance. Notably, UGainS demonstrates a worse FPR95 accuracy on the Road Anomaly dataset compared to RbA when training with OoD data.
% \textcolor{red}{Aug 2023}

To improve robustness to noisy anomaly maps, Zhao et al. propose Score-to-Mask \textbf{S2M} \cite{zhao2024segment}, which utilizes box prompts instead of point prompts. Point prompts can be influenced by false positives even when these regions do not resemble real objects. S2M can be built on top of other OoD detectors that generate anomaly score maps, such as RPL \cite{liu2023residual} or PEBAL \cite{tian2022pixel}. The S2M prompt generator processes the anomaly score maps and produces box prompts for out-of-distribution objects. It is implemented as an object detection module and trained on an outlier exposure dataset. SAM is frozen, and only the prompt generator needs to be updated during training. During inference, box prompts and the original image are fed to SAM, which produces precise masks for OoD objects. S2M achieves strong results on SAM-B, demonstrating its effectiveness and promise for automated driving systems. However, the prompt generator may overlook extremely small road obstacles which are not noticed by OoD detector.
% \textcolor{red}{Nov 2023}

Shoeb et al. propose a framework that leverages features from the SAM decoder and likelihood ratio (\textbf{LR}) to predict out-of-distribution objects \cite{shoeb2025segmentlevelroadobstacledetection}. It eliminates the need for an OoD detector, but results in suboptimal performance on pixel-level OoD segmentation. To explore human interaction in OoD segmentation, Shoeb et al. \cite{shoeb2024have} introduce text queries into their pipeline, which we abbreviate as \textbf{ROD}. They use RbA score to get OoD object masks in individual frames. Then, they link similar segments across subsequent frames using a lightweight object tracker. Finally, both the cropped OoD road obstacle patches and user-provided text queries are embedded using CLIP to retrieve relevant OoD objects based on the users input. A fundamental challenge of this method is that users may not always know how to describe unseen objects, as any novel entity could appear on the road.

\begin{table}[ht]
\centering
\footnotesize
\resizebox{\linewidth}{!}{%
\setlength{\tabcolsep}{3pt}
\begin{tabular}{@{}lcccccc@{}}
\toprule
\textbf{Method} & \textbf{Powerful model} & \textbf{OoD detector} & \textbf{Fine-tuning} & \textbf{Pipeline} & \textbf{OoD info source} & \textbf{Key idea}\\
\midrule
Mask2Anomaly~\cite{rai2023unmasking} & Mask2Former & \xmark & \cmark & B & High-entropy background masks & Mask-classification reformulation; global mask attention \\[2pt]
RbA~\cite{nayal2023rba}              & Mask2Former & \xmark & \xmark & B & Rejected-by-all class prob. & Aggregate class probabilities to score masks without retraining \\[2pt]
EAM~\cite{grcic2023advantages}       & Mask2Former & \xmark & \cmark &B & Low-confidence masks + neg. sup. & Uncertainty aggregation across masks with negative supervision \\[2pt]
UNO~\cite{delic2024outlier}          & Mask2Former & \xmark & \cmark & B& Uncertainty + neg. object logit & Decouple uncertainty and anomaly via UNO score \\[2pt]
CSL~\cite{zhang2024csl}              & Mask2Former & \xmark & \cmark or \xmark & C & Teacher masks / region proposals & Structural constraints / mask splitting to reduce bias \\[2pt]
\midrule
UGainS~\cite{nekrasov2023ugains}     & SAM & \cmark & \xmark & A & RbA anomaly map & Point-prompt from anomaly map \\[2pt]
S2M~\cite{zhao2024segment}           & SAM & \cmark & \xmark & A & RPL/PEBAL anomaly map & Box-prompt generator for noise filtering \\[2pt]
LR~\cite{shoeb2025segmentlevelroadobstacledetection} & SAM & \xmark & \xmark & C & SAM decoder features & Likelihood-ratio on SAM features \\[2pt]
ROD~\cite{shoeb2024have}             & CLIP & \cmark & \xmark & B & RbA masks + CLIP text embed. & Retrieve OoD objects via user text/ semantic similarity \\[2pt]
PixOOD~\cite{vojivr2024pixood}       & DINOv2 & \xmark & \xmark & B & DINOv2 features & Probabilistic modelling in 2-D projected space \\ 
\bottomrule
\end{tabular}}
\caption{Overview of OoD-segmentation methods that leverage powerful models.
The \textbf{Fine-tuning} column indicates whether the powerful model is fine-tuned (\cmark) or frozen (\xmark), and the \textbf{OoD info source} column specifies the origin of the anomaly cues.}
\label{tab:powerful_models}
\end{table}

While training on an OE dataset can enhance OoD segmentation, it may introduce certain biases. \textbf{PixOOD} \cite{vojivr2024pixood} proposes a framework for OoD segmentation that avoids training on anomalous data. It uses a frozen DINOv2 \cite{oquab2024dinov2learningrobustvisual} to extract features, which are projected into 2D space. For each class, PixOOD models the in-distribution and out-of-distribution likelihoods and defines the OoD score. Compared to GROOD \cite{vojivr2023calibrated}, PixOOD supports pixel-level segmentation by upscaling the feature maps at inference. This work demonstrates that OoD segmentation can be achieved using general-purpose visual features. The approach is general since it eliminates the need for training on anomalous data.

Table \ref{tab:powerful_models} summarizes the OoD segmentation approaches leveraging powerful pretrained models, detailing their backbone, pipeline type, fine-tuning strategy, anomaly cue sources, and core ideas. Although these methods have demonstrated impressive performance, the increased number of parameters also leads to higher computational costs and energy consumption. Further research is needed to assess the feasibility of deploying such large models within the limited computational resources of onboard systems. Additionally, their inference time should be evaluated in real-time application scenarios.

\begin{table*}[ht]
\centering
\begin{tabular}{llcccc}
\toprule
\multirow{2}{*}{Category} & \multirow{2}{*}{Method} & \multicolumn{2}{c}{SMIYC-Anomaly} & \multicolumn{2}{c}{SMIYC-Obstacle} \\
\cmidrule(r){3-4} \cmidrule(r){5-6}
& & FPR $\downarrow$ & mean F1 $\uparrow$ & FPR $\downarrow$ & mean F1 $\uparrow$ \\
\midrule
\multirow{4}{*}{Test-time} 
& MSP \cite{hendrycks2016baseline}           & 72.02 & 5.37 & 16.6   & 6.25 \\ %Mask2anomaly

& SML \cite{jung2021standardized}           & 39.5  & 12.20 & 36.8  & 3.00 \\%Mask2anomaly
& ODIN \cite{liang2017enhancing}          & 71.68  & 5.15  & 15.28  & 9.37  \\
& Mahalanobis \cite{lee2018simple}  & 86.99  & 2.68 & 13.08  & 4.70 \\ %Mask2anomaly
\midrule
\multirow{4}{*}{Outlier Exposure}
& RPL \cite{liu2023residual}           & 7.18  & 30.16 & 0.09   & 56.69 \\ %pixelood+RPL
& PEBAL  \cite{tian2022pixel}       & 40.82  & 14.48 & 12.68  & 5.54  \\%pixelood
& Meta-OoD \cite{chan2021entropy}     & 15.00  & 28.72    & 0.75   & 48.51    \\ %Mask2anomaly
& DenseHybrid \cite{grcic2022densehybrid}  & 9.81   & 31.08 & 0.24   & 50.72 \\%pixelood
\midrule
\multirow{4}{*}{Reconstruction}
& DUIR \cite{lis2019detecting}         & 25.93  & 12.51 & 4.70   & 8.38  \\%pixelood
& JSR-Net \cite{vojir2021road}       & 43.85  & 13.66 & 28.86  & 11.02 \\%pixelood
& DaCUP \cite{vojivr2023image}         & --     & --    & 1.13   & 46.01 \\ %pixelood
& SynBoost \cite{di2021pixel}      & 61.86  & 9.99 & 3.15   & 37.57 \\ %pixelood
\midrule
\multirow{8}{*}{Powerful Models}
& RbA \cite{nayal2023rba}           & 11.60  & 46.80 & 0.50   & 60.90 \\ %source RbA
& EAM \cite{grcic2023advantages}           & 4.09   & \textbf{60.86} & 0.52   & 75.58 \\%pixelood ram无
% & RbA \cite{nayal2023rba}           & 4.60   & 51.87 & 0.08   & 57.44 \\ %pixelood  rba无
% & Mask2Anomaly \cite{rai2023unmasking}   & 14.63  & 47.16 & 0.20   & 68.15 \\%pixelood
& Mask2Anomaly \cite{rai2023unmasking}   & 14.60  & 48.60 & 0.20   & 69.80 \\%Mask2anomaly
& PixOOD \cite{vojivr2024pixood}       & 54.33  & 19.82 & 0.30   & 50.82 \\%pixelood的两个都没问题
& UNO \cite{delic2024outlier}           & 2.00   & --    & 0.16   & 77.65 \\ %uno主表好像是test set，不合适
& CSL \cite{zhang2024csl}               &7.16    &50.39   &0.67   &51.02   \\%pixelood
& LR \cite{shoeb2025segmentlevelroadobstacledetection}            & --     & --    & 0.20   & \textbf{78.40} \\
& S2M \cite{zhao2024segment}           & \textbf{1.04} & 60.4 & \textbf{0.02} & 64.96 \\
\bottomrule
\end{tabular}
\caption{Comparison of OoD Segmentation Methods on SMIYC-Anomaly and SMIYC-Obstacle. Lower FPR and higher mean F1 are better.}
\label{tab:comparison}
\end{table*}

\section{Future of OoD Segmentation}
\label{sec:conclusion}
Recent advances in OoD segmentation have made it practical to accurately segment OoD objects at the pixel level. This achievement can provide safety guidance to the perception module in automated driving systems, enabling it to operate more cautiously or request human takeover when necessary. However, several challenges remain and should be explored in future research.

% Define a command for the bold X (adjust font if needed)
\newcommand{\boldx}{\textbf{X}}
% Define checkmark
% \newcommand{\cmark}{\checkmark}

% paper can be added
% Segmenting Objectiveness and Task-awareness Unknown Region for Autonomous Driving
% Diffusion for Out-of-Distribution Detection on Road Scenes and Beyond
% Open-set Anomaly Segmentation in Complex Scenarios

\textbf{Build large-scale, realistic benchmarks for pixel-level OoD segmentation}. 
Existing research mainly evaluates methods on small datasets, which do not fully reflect real driving scenarios. Compared with large benchmarks in other communities, thousands of images are relatively limited for evaluating diverse approaches.
  
\textbf{Design segmentation methods that maintain performance of OoD segmentation on diverse environmental conditions.}
In practice, semantic shifts (OoD object) often occur with covariate shift (noisy such as rain, snow) at the same time. These covariate shifts further increase the difficulty of OoD segmentation. Recent works have begun to leverage vision foundation models to tackle these inputs \cite{keser2025benchmarking}. Some others try to solve it by test-time adaptation methods \cite{gao2024unifiedentropyoptimizationopenset, lee2023towards, li2023robustness, yuan2023robust, yu2024stamp}. However, these approaches focus on image‐level detection rather than per‐pixel segmentation in adverse environments. More effort is needed to achieve OoD segmentation with covariate-shifted input.

\textbf{Leverage off-the-shelf vision foundation models for pixel-level OoD segmentation.}
PixOOD \cite{vojivr2024pixood} demonstrates the effectiveness of using off-the-shelf vision foundation models for OoD segmentation. This approach shows great promise, especially with the rapid advancement of increasingly powerful vision foundation models.

\textbf{Incorporate natural language in task-specific OoD segmentation.}
It is to incorporate natural language in OoD segmentation. Although it may seem paradoxical to describe an unseen object, we can describe the in-distribution environment. For example, urban driving versus off-road implies different driving environments, and personal vehicle versus police car in pursuit reflects different contexts. By providing such language-based priors, we can tailor OoD segmentation models to task-specific environments and improve their performance in different settings.
  % \item 生成数据训练，难点在于图像生成的标签。

% \begin{acks}
% \end{acks}

\bibliographystyle{ACM-Reference-Format}
\bibliography{sample-base}

\appendix

\end{document}